\documentclass[runningheads]{llncs}
\usepackage[T1]{fontenc}
\usepackage{graphicx}
\usepackage{tabularx, multirow, float}
\usepackage{placeins}
\usepackage{csquotes}
\usepackage{eurosym}

\usepackage{url}
\usepackage{ifthen}
\usepackage{url}
\usepackage{breakurl}
\usepackage[super]{nth}
\usepackage[breaklinks]{hyperref}
\usepackage{advdate}

\usepackage{siunitx}
\usepackage{soul}

\usepackage{graphicx}

\usepackage[labelformat=simple]{subcaption}

\usepackage{multirow}

\usepackage{cleveref}
\usepackage{booktabs}
\usepackage{multirow}

\usepackage{svg}

\usepackage{xcolor}
\usepackage{mdframed}

\definecolor{siemens}{HTML}{ec6602}

\usepackage[misc]{ifsym}

\newcommand{\cancerscout}{CancerScout }

\newcommand{\new}[1]{\textcolor{blue}{#1}}

\newcommand{\siemenstxt}[1]{\textcolor{siemens}{#1}}

\renewcommand{\thesubfigure}{(\alph{subfigure})}

\renewcommand{\new}[1]{#1}

\begin{document}

\title{End-to-end Learning for Image-based Detection of Molecular Alterations in Digital Pathology} % \thanks{Supported by organization x.}}

\titlerunning{End-to-end Learning in Digital Pathology}

% \author{Anonymous Authors}

\author{Marvin Teichmann \inst{1}\textsuperscript{(\Letter)} \and Andre Aichert \inst{1} \and Hanibal Bohnenberger \inst{2} \and Philipp Ströbel \inst{2} \and Tobias Heimann \inst{1}}
% index{Teichmann, Marvin} 
% index{Aichert, Andre} 
% index{Bohnenberger, Hanibal} 
% index{Ströbel, Philipp} 
% index{Heimann, Tobias} 

\authorrunning{M. Teichmann et al.}

 \institute{Digital Technology and Innovation, Siemens Healthineers, Erlangen, Germany 
 % \footnote[1]{\email{\{marvin.teichmann, andre.aichert, tobias.heimann\}@siemens-healthineers.com}}
\email{marvin.teichmann@siemens-healthineers.com} 
% \email{andre.aichert@siemens-healthineers.com} 
% \email{tobias.heimann@siemens-healthineers.com} 
\and Institute of Pathology, University Medical Center Goettingen, Goettingen, Germany
 % \email{hanibal.bohnenberger@med.uni-goettingen.de} \\
 % \email{philipp.stroebel@med.uni-goettingen.de}
 % \footnote[4]{\email{\{hanibal.bohnenberger, philipp.stroebel\}@med.uni-goettingen.de}} \\
 }

%     Springer Heidelberg, Tiergartenstr. 17, 69121 Heidelberg, Germany
%     \email{lncs@springer.com}\\
%     \url{http://www.springer.com/gp/computer-science/lncs} \and
%     ABC Institute, Rupert-Karls-University Heidelberg, Heidelberg, Germany\\
%     \email{\{abc,lncs\}@uni-heidelberg.de}}
%
\maketitle              % typeset the header of the contribution
\begin{abstract}
    Current approaches for classification of whole slide images (WSI) in digital pathology predominantly utilize a two-stage learning pipeline. The first stage identifies areas of interest (e.g. tumor tissue), while the second stage processes cropped tiles from these areas in a supervised fashion. During inference, a large number of tiles are combined into a unified prediction for the entire slide. A major drawback of such approaches is the requirement for task-specific auxiliary labels which are not acquired in clinical routine. 
    
    We propose a novel learning pipeline for WSI classification that is trainable end-to-end and does not require any auxiliary annotations. We apply our approach to predict molecular alterations for a number of different use-cases, including detection of microsatellite instability in colorectal tumors and prediction of specific mutations for colon, lung, and breast cancer cases from The Cancer Genome Atlas. Results reach AUC scores of up to 94\% and are shown to be competitive with state of the art two-stage pipelines. We believe our approach can facilitate future research in digital pathology and contribute to solve a large range of problems around the prediction of cancer phenotypes, hopefully enabling personalized therapies for more patients in future.
    
\end{abstract}

% \section{Introduction}

%    Main Goals:
%
%   * describe setting and why important
%
%   * describe two-stage approaches 
%       * includes related work / relevant paper
%
%   * name drawback and issues with two-stage approaches
%       * auxiliar annotations
%       * force prediction for every tile (information may not be present)
%
%   * highlight our contribution

\begin{figure}[t]
\centering

\includegraphics[width=\textwidth]{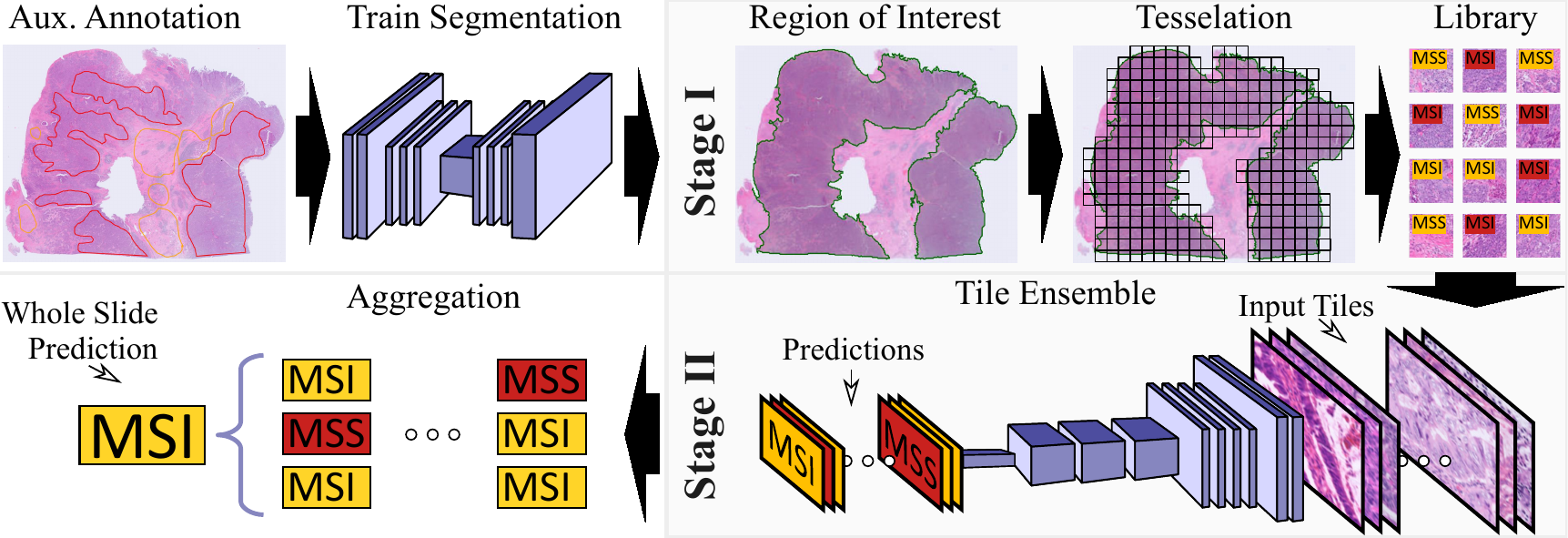}

\caption{Visualization of a typical two stage pipeline.}
% \vspace{-0.35cm}
\label{fig:overview_two_stage}
\end{figure}

As Whole Slide Imaging (WSI) is becoming a common modality in digital pathology, large numbers of highly-resolved microscopic images are readily available for analysis. Meanwhile, precision medicine allows for a targeted therapy of more and more cancer types, making the detection of actionable genetic alterations increasingly valuable for treatment planning and prognosis. Over the last few years, several studies have focused on the prediction of specific mutations, molecular subgroups or patient outcome from microscopy data of tumor tissue \cite{echle2021deep,coudray2018classification,kather2020pan}. The large size of WSI images and the localized nature of information have led to the development of specific processing pipelines for this application.

In a comprehensive review, Echele et al. \cite{echle2021deep} observe that the majority of work on WSI classification comprises two stages. Depending on the task at hand, the first stage selects a region of interest (ROI) of a certain type of tissue or high tumor content \cite{kather2020pan,echle2020clinical,lee2021two}, while some tasks \cite{nagpal2019development,strom1907pathologist} and methods \cite{fu2020pan,nazeri2018two} require even more detailed localized annotation. This stage typically involves a separately trained segmentation model. 
In the second stage, tessellation of the ROI creates a set of smaller tiles (e.g. $224 \times 244$ pixels) that are well suited for processing with convolution neural networks (CNNs). For training, each tile is assigned the same target label corresponding to the whole slide. During inference, a subset or all of the tiles from a ROI are classified by the CNN. In order to obtain a slide-level prediction, all tile-level predictions are combined, e.g. by averaging the confidences \cite{kather2020pan}, class voting \cite{coudray2018classification} or by a second-level classifier \cite{nagpal2019development}. We visualize a typical two-stage pipeline in \Cref{fig:overview_two_stage}. \new{Some studies \cite{echle2021deep,laleh2021benchmarking} omit the segmentation step and randomly choose tiles across the entire slide. This adds label noise to the classification step, since some areas (e.g. healthy tissue) do not contain any relevant information for the classification task at hand, which decreases the prediction performance.}

Recently, a few works which avoid auxiliary annotations have been presented.
Weakly supervised methods aim to implicitly identify tiles with high information value without manual annotation \cite{campanella2019clinical,chen2019rectified,hou2016patch}. % In the space of end-to-end learning, Hemati at al. \cite{hemati2021cnn} proposes a method based on permutation-invariant vector embeddings for representation learning.
In another line of work, clustering-based methods have been proposed for end-to-end WSI classification \cite{lu2021data,shao2021transmil,sharma2021cluster}. \new{A recent benchmark \cite{laleh2021benchmarking} compares a number of state-of-the-art weakly supervised and end-to-end training methods for WSI classification. Their results indicate that the known weakly supervised and end-to-end methods are unable to outperform the widely used two-stage prediction pipeline. The existing methods therefore effectively trade annotation effort for prediction performance.}

% A recent benchmark ... An overview

In this paper, we introduce a $k$-Siamese CNN architecture for WSI classification which is trainable end-to-end, does not require any auxiliary annotations, and is straight-forward to implement. We show that our method outperforms a reference two-stage approach in the clinically relevant task of microsatellite instability (MSI) classification in WSI of formalin-fixed paraffin-embedded (FFPE) slides with haematoxylin and eosin (H\&E) stained tissue samples of colorectal cancer.
In addition, we present competitive results on multiple tasks derived from a range of molecular alterations for breast, colon and lung cancer on the public Cancer Genome Atlas database (TCGA).

\section{Our Method: $k$-Siamese Networks}

\label{sec:e2e}

\new{We believe that the main reason for the success of two-stage approaches is that they mitigate the label noise issue inherent to tile based processing.} Training a classifier on every tile from a WSI separately is disadvantageous since a large number of tiles do not contain any visual clues on the task at hand. Tiles showing only healthy tissue for example do not contain any information about the tumor. We know that CNNs are able to overfit most datasets, if this is the optimal strategy to minimize training error \cite{zhang2017understanding}. Utilizing uninformative tiles during training therefore results in the network learning features which degrade its generalization ability. We believe that this is the main reason that led to two-stage approaches becoming so popular for WSI analysis. \new{However, for some tasks only a subset of the tumor area might contain the relevant information, for other tasks it might be required to combine visual information from multiple tiles before taking a decision. Both scenarios are not handled well by current two-stage pipelines.}

\begin{figure*}[t]
    \centering

    \includegraphics[width=\textwidth]{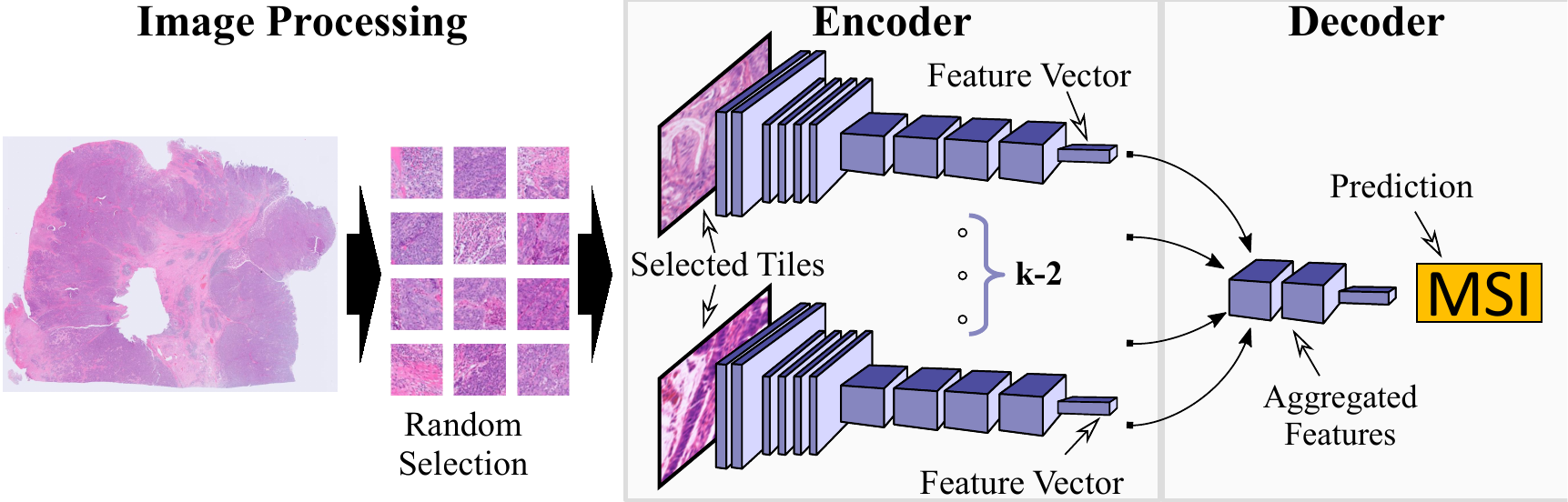}
    % \begin{mdframed}[backgroundcolor=black]
    % \begin{subfigure}[t]{0.30\textwidth}
    %     \centering
    %     \includegraphics[height=1.25in]{figs/preprocess_orange.png}
    %     \caption{\siemenstxt{Image Preprocessing}}
    %     \label{fig:prepross}
    % \end{subfigure}%
    % ~ 
    % \begin{subfigure}[t]{0.8\textwidth}
    %     \centering
    %     \includegraphics[height=1.25in]{figs/network_white_orange.png}
    %     \caption{\siemenstxt{$k$-Siamese Network Architecture}}
    %     \label{fig:arch}
    % \end{subfigure}%
    % \end{mdframed}

    \caption{An overview over our end-to-end learning pipeline.}
    % \vspace{-0.1cm}
    \label{fig:onestage}
\end{figure*}

\new{We propose a novel encoder-decoder based pipeline to address these issues.} %aforementioned } 
% \st{We propose an alternative way to mitigate the impact of uninformative input patches: our novel deep learning pipeline is heavily inspired by current two-stage approaches, however, we integrate all learning stages into a unified encoder-decoder  architecture, enabling full end-to-end training.} 
Our encoder produces a latent representation for $k$ randomly selected tiles from the input WSI. These tiles are processed simultaneously while sharing their weights. The resulting set of feature vectors is than aggregated by the decoder to output a single joined prediction. We call our approach $k$-Siamese networks, since it follows the idea of Siamese networks, but with $k$ instead of just two encoders. We illustrate our approach in \Cref{fig:onestage}.

\new{The feature vectors produced by the encoder are learned implicitly and can store any kind of information, including that the tile is not meaningful for the task at hand. The decoder can learn to interpret those feature vectors and combine the information found in multiple tiles.} If $k$ is chosen large enough, a sufficient number of the selected tiles contain task-relevant information, which eliminates the need for any auxiliary annotations.

\renewcommand{\thesubfigure}{\siemenstxt{(\alph{subfigure})}}

\subsubsection{Design Choices}

Our encoder is based on Efficientnet-B0 \cite{tan2019efficientnet}, which offers high predictive capability with a relatively small computational and memory footprint. Our decoder performs average pooling over the feature vectors from all $k$ patches, followed by a $1 \times 1$ convolution and a softmax layer.  We have evaluated more complex designs however, we did not observe any significant performance boost. Utilizing adaptive average pooling for the feature vector aggregation step has the additional benefit that the model can be used with a variable number of encoders. This allows us to perform memory efficient training with as few as \num{24} tiles, while using more tiles for better prediction performance during inference.

\subsubsection{Training and Inference}

\label{sec:training}

Our model is trained with stochastic gradient-descent using the Adam heuristic \cite{kingma2014adam}. For training the encoder, we use a fine-tuning approach and start with the official EfficientNet weights, provided Tan et al. \cite{tan2019efficientnet}. Unless otherwise specified, we use the following training parameters for all our experiments: base learning-rate ($blr$) of \num{2e-5} and batch-size ($bs$) of \num{6}. 

% we use the default hyper-parameters given in \Cref{tab:hs} for all our experiments.

% 
Following the discussions in \cite{goyal2017accurate}, we normalize our learning-rate ($nlr$) by multiplying the base-learning rate ($blr$) with our batch-size ($bs$): $nlr = bs \times blr$. We train the model for \num{72} epochs and report the scores evaluated on the final epoch. We use \num{12} warm-up epochs, during which the learning rate ($lr$) is linearly increased from \num{0} to $nlr$ \cite{goyal2017accurate}. For the remaining \num{60} epochs, we use polynomial learning rate decay \cite{liu2015parsenet}. We use automatic mixed precision (amp) \cite{micikevicius2017mixed} training to reduce the memory and computational footprint. To improve generalization, we use the following regularization methods: We apply quadratic weight decay with a factor of \num{5e-4} to all our weights. We use dropout \cite{srivastava2014dropout} for the decoder and stochastic depth \cite{huang2016deep} for the encoder. We apply data-augmentation to each tile independently. We use the following common data-augmentation methods: (random) brightness, contrast, saturation, hue and rotation. In addition, tiles are not taken from a fixed grid, but their location is chosen randomly but non-overlapping. We exclude tiles which only contain background, which is estimated by using a threshold on the colour values.

% following training parameters for all our experiments. We use a base learning-rate ($blr$) of \num{2e-5} and batch-size ($bs$) of \num{6}. 

During training, we use \num{24} tiles per slide, each with a spatial resolution of $256 \times 256$ pixel. We perform inference on \num{96} tiles. All tiles have an isometric resolution of \num{0.25} microns/pixel, which corresponds to a $10\times$ optical magnification.

\section{Data}

\renewcommand{\thesubfigure}{{(\alph{subfigure})}}
\begin{figure*}[t]
    \centering
    % \begin{mdframed}[backgroundcolor=black]
    % \vspace{-0.2cm}
    \begin{subfigure}[t]{0.33\textwidth}
        \centering
        \includegraphics[height=1.2in]{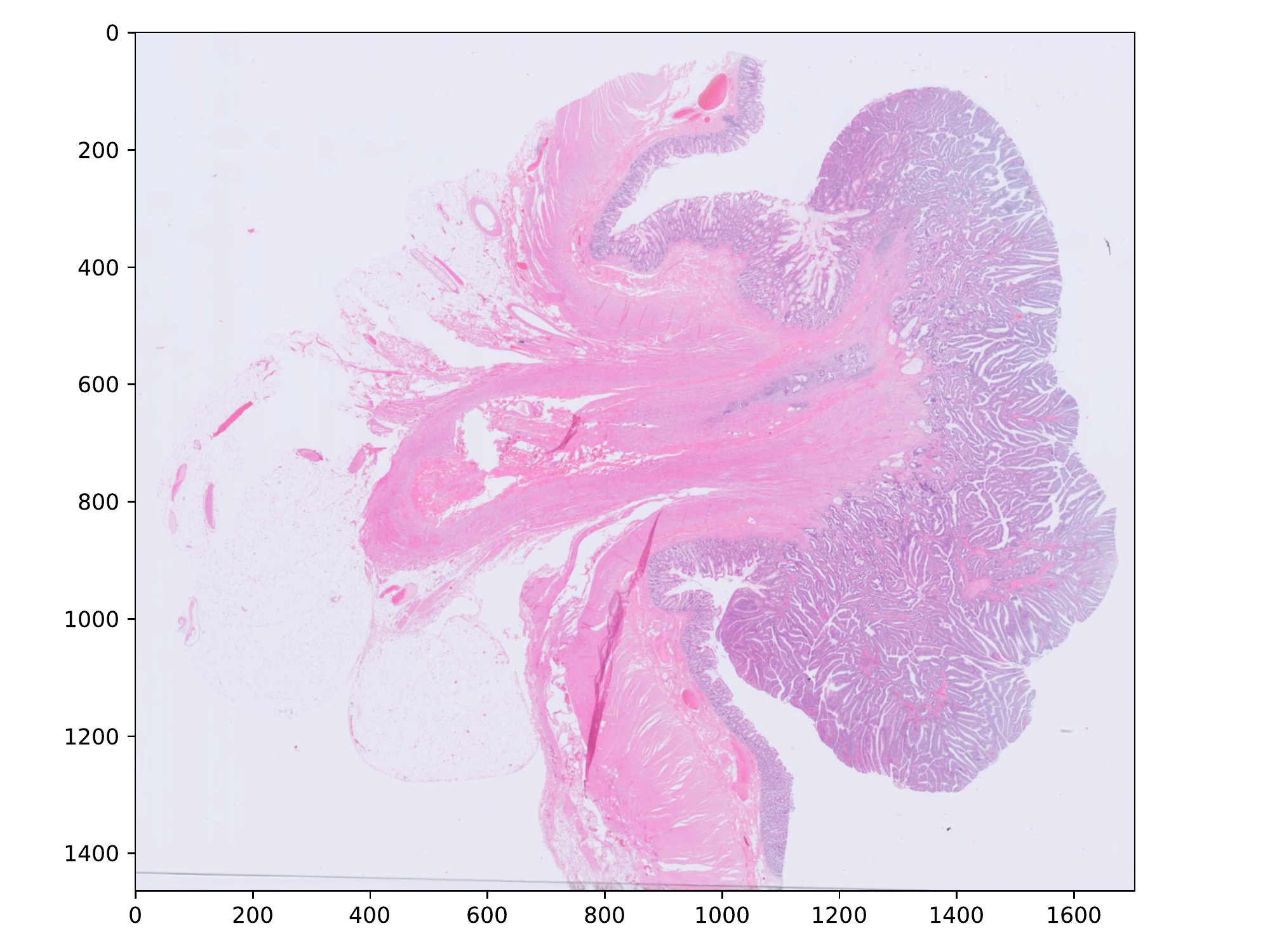}
        \caption{MSI negative}
        \label{fig:prepross}
    \end{subfigure}%
    ~
    \begin{subfigure}[t]{0.33\textwidth}
        \centering
        \includegraphics[height=1.2in]{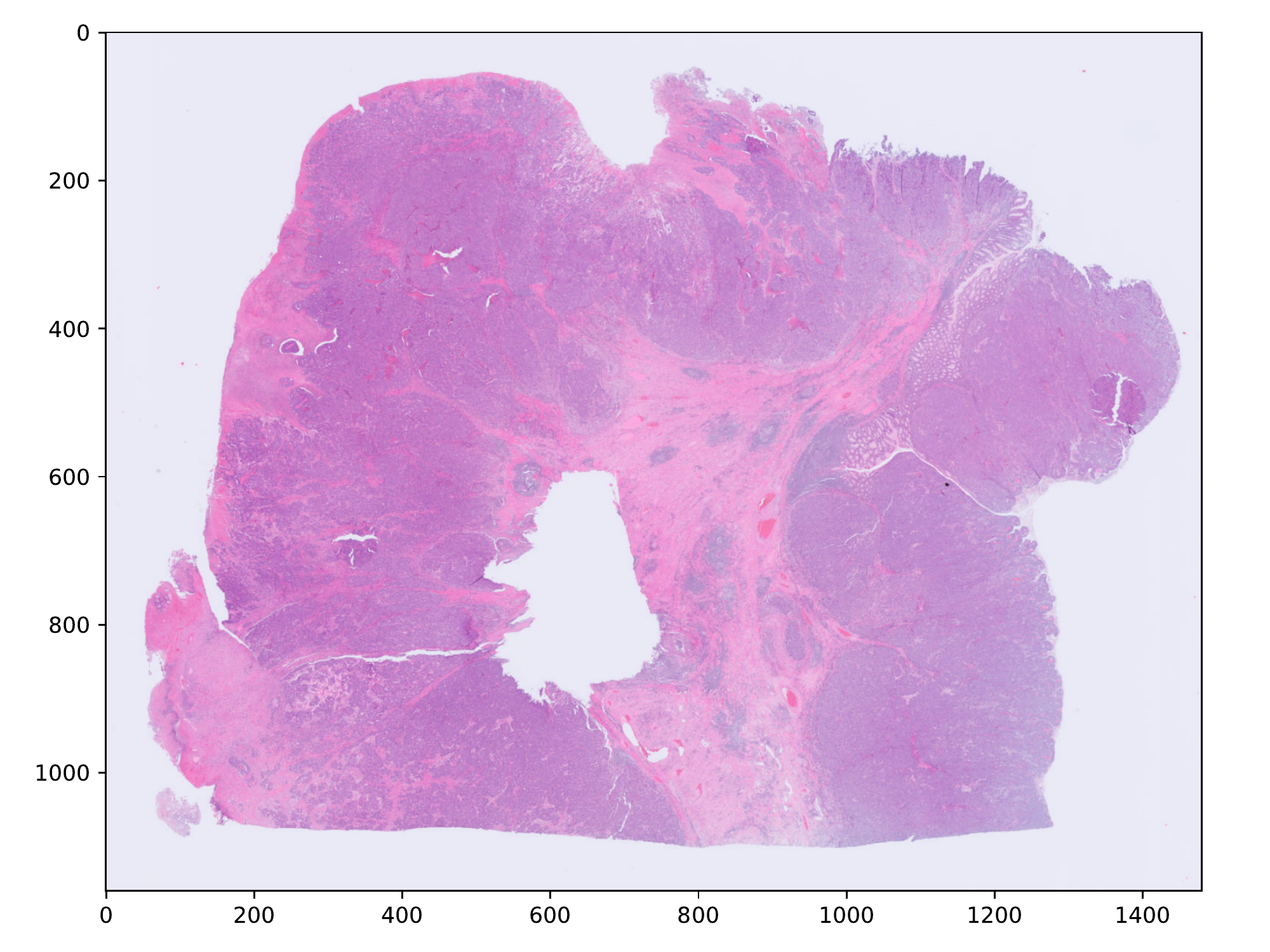}
        \caption{MSI positive}
        \label{fig:arch}
    \end{subfigure}%
    ~ 
    \begin{subfigure}[t]{0.33\textwidth}
        \centering
        \includegraphics[height=1.2in]{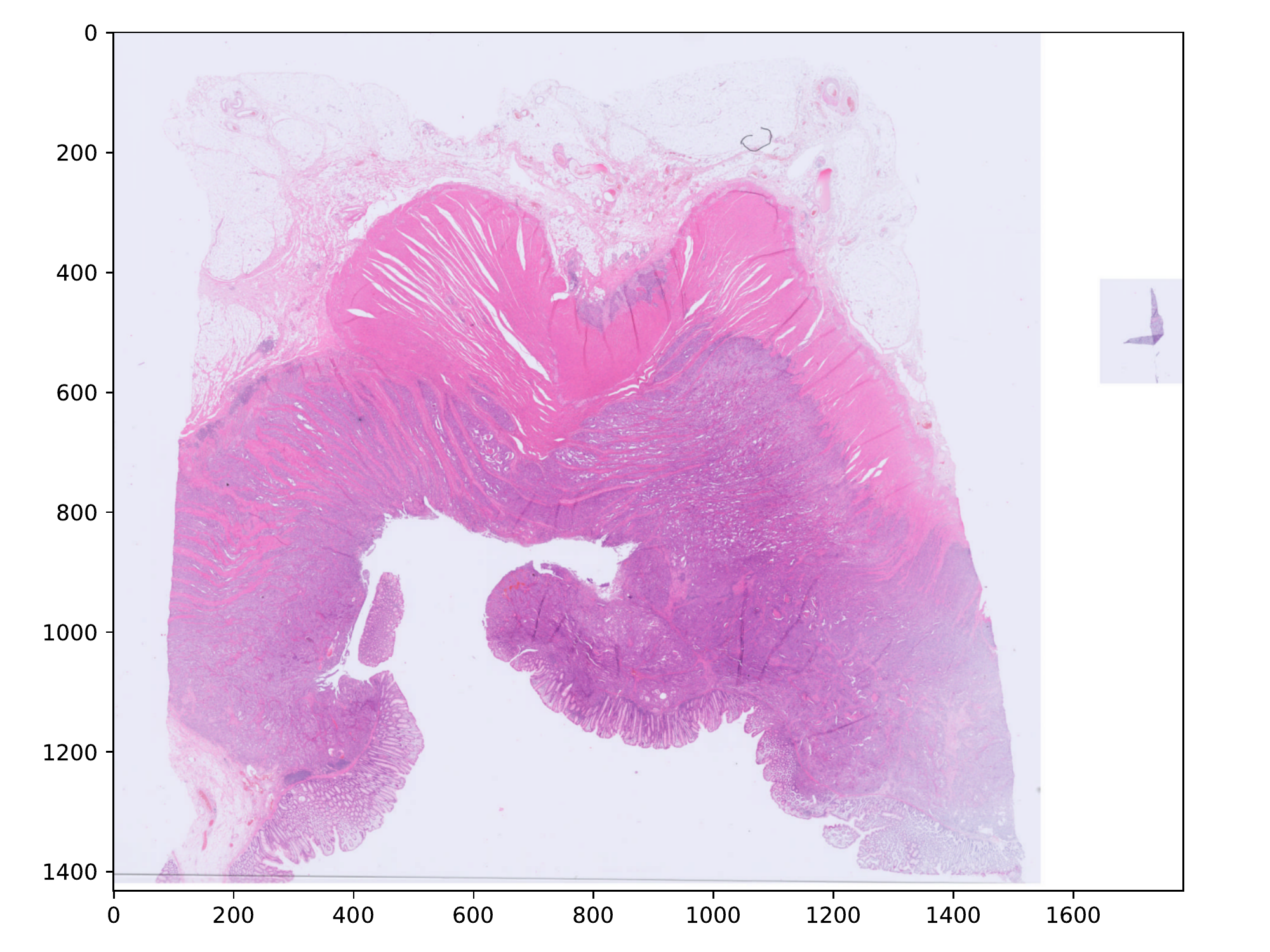}
        \caption{MSI negative}
        \label{fig:arch}
    \end{subfigure}%

    \caption{Three examples of diagnostic slides from the \cancerscout Colon dataset. Slides are plotted with an optical magnification of 2.5.}
    % \vspace{-0.3cm}
    \label{fig:ex_imgs}
\end{figure*}

\subsection{The \cancerscout Colon Data}

For this study, we use \num{2085} diagnostic slides from \num{840} colon cancer patients. We have estimated the MSI status of all patients using clinic immunohistochemistry (IHC) based test. A total of \num{144} (\SI{17}{\percent}) patients in the cohort are MSI positive. In addition, we have annotated tumor regions in \num{299} slides from \num{279} patients, with the open-source annotation tool EXACT \cite{marzahl2021exact}. We use these annotations to train a segmentation model for our reference two-stage approach. Ethics approval has been granted by \new{University Medical Center Goettingen (UMG)}.

% \renewcommand{\thesubfigure}{{(\alph{subfigure})}}
% \begin{figure*}[t]
%     \centering
%     % \begin{mdframed}[backgroundcolor=black]
%     % \vspace{-0.2cm}
%     \begin{subfigure}[t]{0.18\textwidth}
%         \centering
%         \includegraphics[height=0.75in]{figs/cnew_plot.png}
%         \caption{cnew}
%         \label{fig:prepross}
%     \end{subfigure}%
%     ~
%     \begin{subfigure}[t]{0.32\textwidth}
%         \centering
%         \includegraphics[height=0.75in]{figs/cold_plot.png}
%         \caption{cold}
%         \label{fig:arch}
%     \end{subfigure}%
%     ~ 
%     \begin{subfigure}[t]{0.18\textwidth}
%         \centering
%         \includegraphics[height=0.75in]{figs/hnew_plot.png}
%         \caption{hnew}
%         \label{fig:arch}
%     \end{subfigure}%
%     ~ 
%     \begin{subfigure}[t]{0.32\textwidth}
%         \centering
%         \includegraphics[height=0.75in]{figs/hold_plot.png}
%         \caption{hold}
%         \label{fig:arch}
%     \end{subfigure}%
%     % \end{mdframed}
% 
%     \caption{The four diagnostic slides of a MSI-negativ patient in the \cancerscout Dataset. Slides are plotted with an optical magnification of \num{1.25}.}
%     \label{fig:ex_sets}
% \end{figure*}

\subsubsection{Patient Cohort}

The patient cohort was defined by \new{pathologist from the UMG} and consist of \num{840} colorectal cancer (CRC) patients. Patients were selected from those treated between 2000 and 2020 at \new{UMG} and who gave consent to be included in medical studies. Only patients with resected and histologically confirmed adenocarcinoma of the colon or rectum were included in this dataset. Among those, the pathologists manually selected samples for which enough formalin-fixed and paraffin embedded tumor tissue for morphological, immunohistochemical and genetic analysis was available. Patients of age 18 or younger and patients with neoadjuvant treatment were excluded from this study. % Neoadjuvant treatment means treatment before surgical resection and excludes most of the rectal cancer patients. Those patients with rectal cancer that have been included did not receive neoadjuvant treatment.

\subsubsection{Image Data} 

The images are magnified H\&E stained histological images of formalin-fixed paraffin-embedded (FFPE) diagnostic slides. Images are scanned with an isometric resolution of \num{0.25} microns/pixel, which corresponds to a microscopic magnification of \SI{40}{\times}. For all patients, a new slide was freshly cut, stained, and digitalized for this study. \Cref{fig:ex_imgs} shows examples of those slides, we call them \textit{cnew} slides. For \num{725} patients we have digitalized \textit{cold} slides. These are archived slides which were cut and stained when the patient was initially treated. Each of the slides is from the same FFPE block as the corresponding \textit{cnew}, located in very close proximity (about \SI{2}{\micro\metre}). Those slides are used to augment training but not for evaluation. For \num{274} patients we have collected \textit{hnew} slides. These are slides which only contain healthy tissue taken from the resection margins of the FFPE block. For \num{246} patients we have collected \textit{hold} slides. These are slides which were cut and stained when the patient was initially treated, located in close proximity (about \SI{2}{\micro\metre}) to the corresponding \textit{hnew} slide % and only contain healthy tissue. 
We use those slides to increase the training data for our segmentation model. % but not for the classification tasks. 

\subsection{TCGA Data}

\label{sec:TCGA}

For additional experiments, we use three datasets based on The Cancer Genome Atlas (TCGA) data. The datasets are designed to perform mutation detection for breast invasive carcinoma, colon adenocarcinoma and lung adenocarcinoma patients and are based on the projects TCGA BRCA \cite{koboldt2012comprehensive}, TCGA COAD\cite{cancer2012comprehensive} and TCGA LUAD \cite{collisson2014comprehensive} respectively. We include all patients of the corresponding projects where the diagnostic slide images were publicly available in January 2022. TCGA diagnostic slides are WSIs from H\&E-stained FFPE tissue of the primary tumor. The image data can be downloaded through the Genomic Data Commons Portal (\url{https://portal.gdc.cancer.gov/}).

We combine the slide images with somatic mutation data which serve as targets. For this, we utilize the omics data computed by the ensemble pipeline proposed in \cite{ellrott2018scalable}. This data can be downloaded using the xenabrowser (\url{https://xenabrowser.net/datapages/}). We only include genes which are considered \textit{Tier 1} cancer drivers according to the Cosmos Cancer Gene Census \cite{sondka2018cosmic}. Of those, we consider the top \num{8} most prevalently mutated genes from each cohort for this study. We consider a gene mutated if it has a non-silent somatic mutation (SNP or INDEL). We exclude all patients from cohorts for which no somatic mutation data are provided. \new{The individual genes, their respective mutation prevalence and} the size of each cohort is given in \Cref{tab:tga_results}.

% The total size of each cohort is given in 

\begin{table}[t]
    \centering
    \begin{minipage}[b]{0.34\textwidth}
    \begin{tabular}{l | r | r }
        \toprule
        Method          & AP & ROC AUC \\[0.4mm]
        \midrule % \\[-13.39mm]
        \midrule \\[-3.39mm]
        Seg-Siam        &   \num{0.83}   &   \num{0.94} \\
        $k$-Siam        &   \num{0.83}   &   \num{0.94} \\
        Two Stage       &   \num{0.77}   &   \num{0.91} \\
        CLAM \cite{lu2021data}           &   \num{0.73}   &   \num{0.90}\\
        ViT \cite{laleh2021benchmarking}             &   \num{0.77}   &   \num{0.89} \\
        MIL \cite{campanella2019clinical}             &   \num{0.69}   &   \num{0.88} \\
        EfficientNet    &   \num{0.69}   &   \num{0.87} \\
        \bottomrule
    \end{tabular}
    \caption{\new{Results on the MSI prediction task (n = 672).}}
    % \vspace{-0.25cm}
    \label{tab:comp}
    \end{minipage}
    \begin{minipage}[b]{0.65\textwidth}
    \begin{tabular}{l | r | r | r | r}
    \toprule
                    & base learning & batch & num & warm up      \\
                    & rate (blr) & size (bs) & epochs & epochs     \\
    \midrule
    range           & [\num{3e-6}, \num{e-4}] & [\num{4}, \num{24}] &  [\num{32}, \num{96}] & [\num{0}, \num{18}] \\
    \midrule
    default         &   \num{2e-5}     &   \num{6}   &   \num{72} & \num{12} \\
    Seg-Siam        &   \num{7.5e-5}   &   \num{19}  &   \num{38} & \num{5} \\
    $k$-Siam        &   \num{5.5e-5}   &   \num{21}  &   \num{81} & \num{16} \\
    Two Stage       &   \num{8.8e-5}   &   \num{12}  &   \num{35} & \num{10} \\
    EfficientNet    &   \num{9.3e-5}   &   \num{21}  &   \num{58} & \num{4} \\
    \bottomrule
    \end{tabular}
    % \caption{The optimal hyperparameters estimated for our models.}
    \caption{Our default as well as the optimal hyperparameters estimated for our models.}
    % \vspace{-0.25cm}
    \label{tab:hs}
    \end{minipage}

\end{table}

\renewcommand{\thesubfigure}{{(\alph{subfigure})}}
\begin{figure*}[t]
    \centering
    % \begin{mdframed}[backgroundcolor=black]
    % \vspace{-0.35cm}
    \hspace{-0.5cm}
    \begin{subfigure}[t]{0.25\textwidth}
        \centering
        \includegraphics[width=\textwidth]{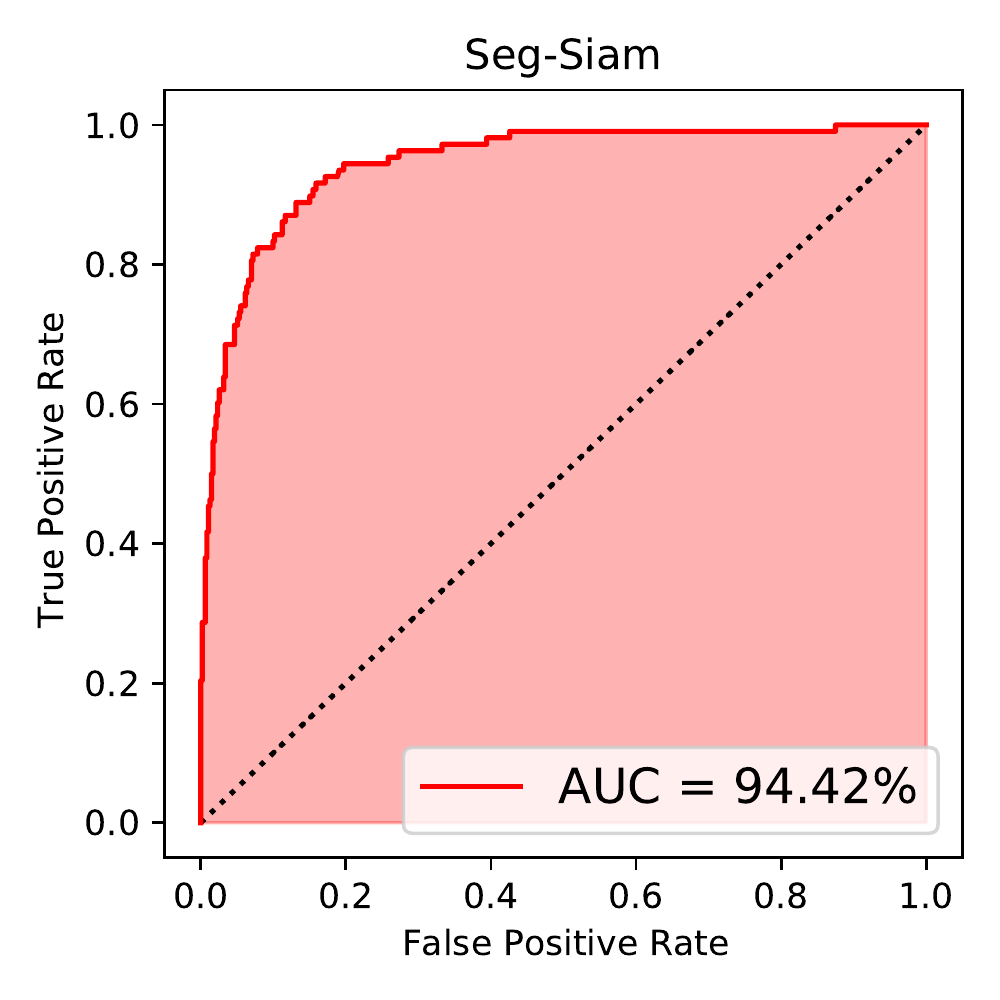}
        % \caption{cold}
        \label{fig:arch}
    \end{subfigure}%
    ~
    \begin{subfigure}[t]{0.25\textwidth}
        \centering
        \includegraphics[width=\textwidth]{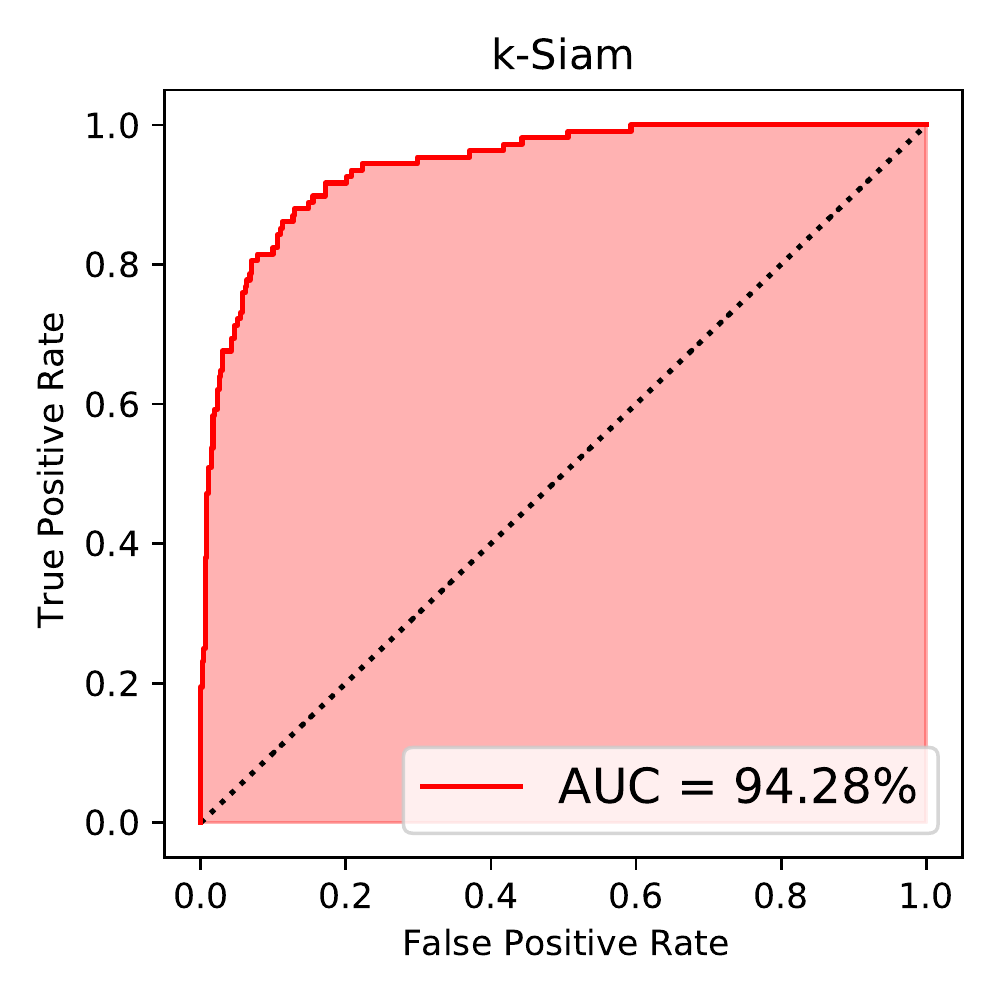}
        % \caption{cnew}
        \label{fig:prepross}
    \end{subfigure}%
    ~ 
    \begin{subfigure}[t]{0.25\textwidth}
        \centering
        \includegraphics[width=\textwidth]{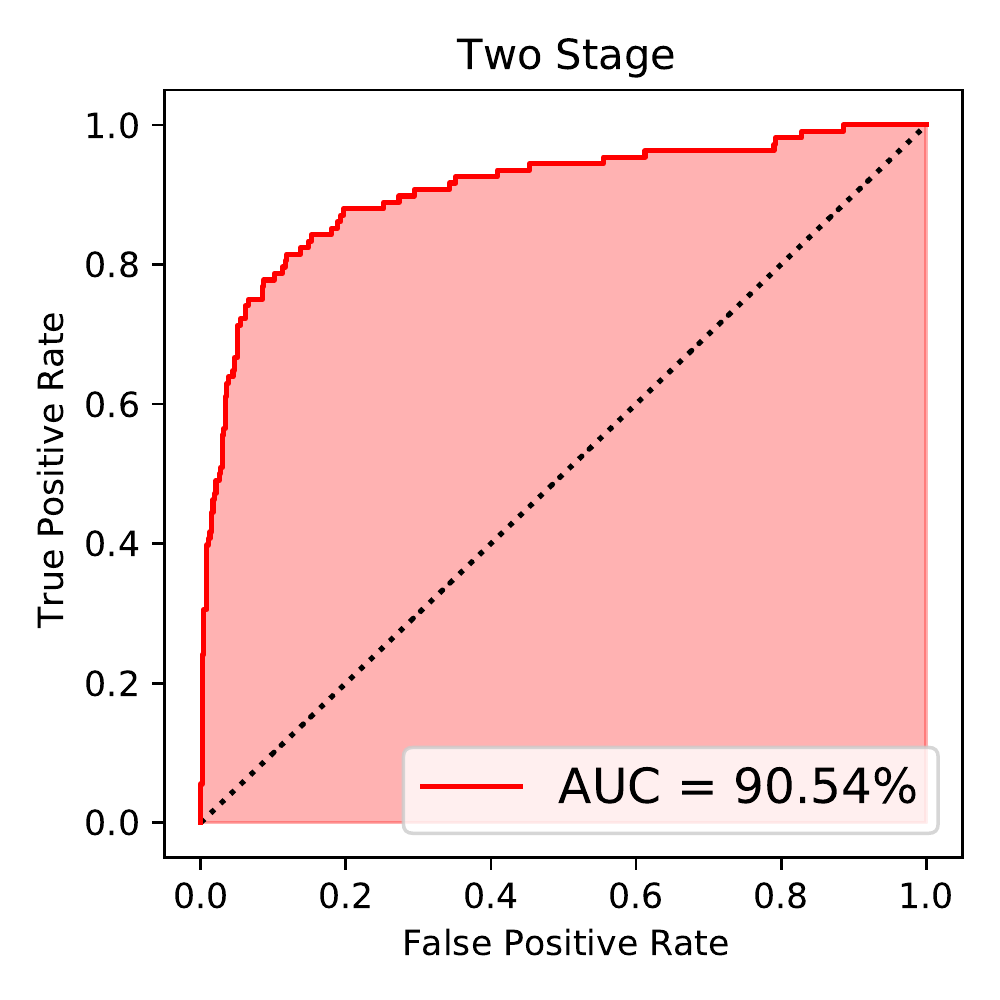}
        \label{fig:prepross}
    \end{subfigure}%
    ~
    \begin{subfigure}[t]{0.25\textwidth}
        \centering
        \includegraphics[width=\textwidth]{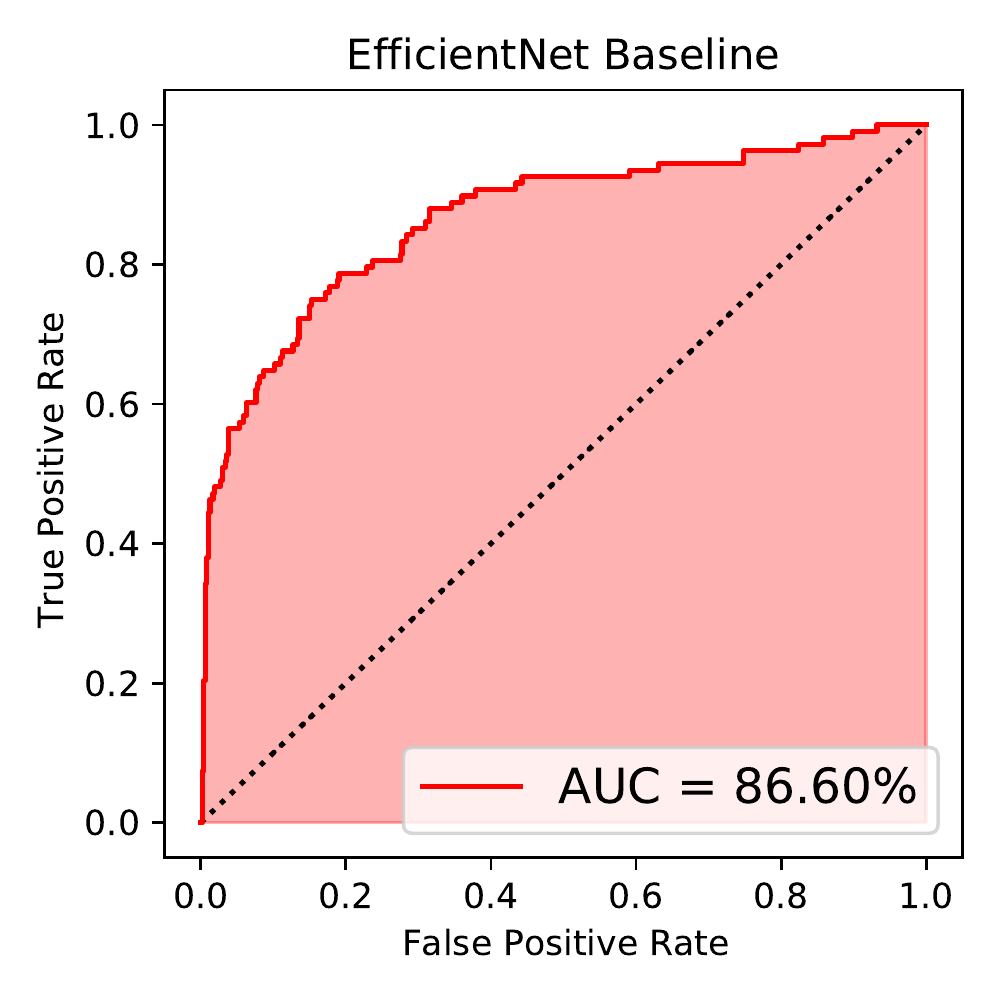}
        % \caption{hnew}
        \label{fig:arch}
    \end{subfigure}%
    % \vspace{-0.5cm}
    % \end{mdframed}
    % \vspace{-0.4cm}
    \caption{RoC curves of the four models on the MSI prediction task (n = 672).}
    % \vspace{-0.15cm}
    \label{fig:abl_roc}
\end{figure*}

\section{Experiments and Results}

\subsection{MSI Prediction}

% \begin{table}[t]
%     \centering
%     \begin{tabular}{l | c | c | c | c}
%         \toprule
%                           & $k$-Siam & Seg-Siam & No Seg Ensemble & Two Stage      \\
%         \midrule
%         Average Precision & \SI{82.91}{\percent} & \SI{83.07}{\percent} &  \SI{68.67}{\percent} & \SI{77.03}{\percent} \\
%         RoC AUC           & \SI{94.28}{\percent} & \SI{94.42}{\percent} &  \SI{86.60}{\percent} & \SI{90.54}{\percent} \\
%         \bottomrule
%     \end{tabular}
% 
%     \caption{Prediction performance on the MSI prediction task (n = -111 Patients).}
%     \label{tab:results}
% \end{table}

We performed an ablation study on the \cancerscout colon data to evaluate the quality and features of our model. In total, we compare the performance of four pipelines in the MSI prediction task. The first, \textit{$k$-Siam}, uses random tile selection followed by the $k$-Siamese network described in \Cref{sec:e2e}. \textit{Seg-Siam} uses tumor segmentation for tile selection followed by a $k$-Siamese network. \textit{Two Stage} uses tumor segmentation for tile selection followed by tile-wise classification, implementing the standard two-stage approach. The \textit{EfficientNet} baseline uses random tile selection and tile-wise classification. RoC curves together with the respective AUC values for all four pipelines are shown in \Cref{fig:abl_roc}. \new{In \Cref{tab:comp} we report the results of our pipelines compared to the methods discussed in \cite{laleh2021benchmarking}.}

% All models are evaluated on a total of \num{288} tiles per slide during inference.

\subsubsection{Experimental Setup}

For the tumor segmentation, we use use a PAN \cite{lipyramid} based model with Efficientnet \cite{tan2019efficientnet} backbone. This approach yields a validation Intersection over Union (IoU) performance of \SI{98}{\percent}. We use Efficientnet-B0 as base-classifier for all our experiments. Prediction aggregation is performed by averaging the confidences of all processed tiles. We use the same training and data-augmentation pipeline for all four models. For a fair comparison, we perform a random hyperparameter search with a total of \num{96} runs per model over the most influential training parameters. The parameters considered, their range and optimal values are given in \Cref{tab:hs}. % All other parameters are chosen identical for all for models.
%TODO Weakly supervised baseline

We evaluate the performance of our models using a \num{5} fold patient-level cross-validation. We use fold \num{0} for the hyperparameter search and folds \num{1} to \num{4} as test-set for evaluation. No parameter-tuning, threshold selection or training decisions are done using those test folds. In particular, we did not do any early stopping based on the evaluation score, but rather train the model to the end and evaluate model performance after the final epoch. 

% We note that our hyperparemeter search indicates that \textit{Seg-Siam} and \textit{Two-Stage} require a lower batch size and fewer epochs. We believe that this is due to the higher information density and reduced noise in the processed data.

% Comparing \textit{Two-Stage} and \textit{No Seg Ensemble} reveals that utilizing segmentation does have a major impact on the model performance for the more traditional pipelines. For such pipelines, collecting auxiliary annotations seems necessary and desirable in order to achieve meaningful results.

% \gray{In addition our hyperparameter search indicates that both tend to require less epochs and a lower batch size for convergence. }

% \subsection{Segmentation}

% our clinical partners provide

% Towards the goal of reducing the annotation efforts for our clinical partners we have designed an annotation protocol requesting incomplete but exact annotations. % The annotator is asked to annotate two kinds of regions in each slides. Firstly diagnostic relevant regions containing dense tumor tissue and secondly regions % containing tissue which does not contain tumor tissue. The latter includes healthy tissue (fat, muscle, lymphatic tissue) 

% The latter includes healthy tissue utilize incomplete

% Two Stage vs One Stage

% \subsection{MSI Detection Performance}

\begin{table}[t]
    \centering
    \hspace{-0.3cm}
    \begin{subtable}[t]{0.33\textwidth}
        \scriptsize
        \begin{tabular}{l | r || r | r }
            \toprule
            \multirow{2}{*}{Gene}   & \multicolumn{1}{c||}{\multirow{2}{0.8cm}{\new{Preva-lance}}}    
            &  \multicolumn{1}{c|}{Ours}      &   \multicolumn{1}{c}{Ref.\cite{kather2020pan}}   \\
             & & \multicolumn{1}{c|}{\new{[AUC]}}   & \multicolumn{1}{c}{\new{[AUC]}}   \\
            \midrule
            PIK3CA & \SI{35}{\percent}   &   \textbf{0.64}  &   \num{0.63} \\
            TP53   & \SI{33}{\percent}   &   \textbf{0.80}  &   \num{0.78} \\
            CDH1   & \SI{13}{\percent}   &   \textbf{0.82}     &   --- \\
            GATA3  & \SI{12}{\percent}   &   \textbf{0.64}     &   <\num{0.62}\\
            KMT2C  & \SI{10}{\percent}   &   \num{0.53}     &   <\num{0.62}\\
            MAP3K1 & \SI{8}{\percent}    &   \num{0.47}     &   \num{0.62} \\
            PTEN   & \SI{6}{\percent}    &   \textbf{0.75}  &   <\num{0.62} \\
            NCOR1  & \SI{5}{\percent}    &   \num{0.51}     &   --- \\
            \bottomrule
        \end{tabular}
        \scriptsize
        \caption{Breast (n = 761)}
        \label{tab:brca}
    \end{subtable}
    ~
    \begin{subtable}[t]{0.32\textwidth}
        \scriptsize
        \begin{tabular}{l | r || r | r }
            \toprule
            \multirow{2}{*}{Gene}   & \multicolumn{1}{c||}{\multirow{2}{0.8cm}{\new{Preva-lance}}}    
            &  \multicolumn{1}{c|}{Ours}      &   \multicolumn{1}{c}{Ref.\cite{kather2020pan}}   \\
             & & \multicolumn{1}{c|}{\new{[AUC]}}   & \multicolumn{1}{c}{\new{[AUC]}}   \\
            % Gene   & ~Prev.~     &  ~Ours~       &   Ref.\cite{kather2020pan}   \\
            \midrule
            APC    &     \SI{74}{\percent}   &   \textbf{0.66}  &   \num{0.65} \\
            TP53   &     \SI{60}{\percent}   &   \textbf{0.74}  &   \num{0.68} \\
            KRAS   &     \SI{42}{\percent}   &   \num{0.63}  &   <\num{0.65} \\
            PIK3CA &     \SI{32}{\percent}   &   \num{0.54}  &   <\num{0.65} \\
            FAT4   &     \SI{27}{\percent}   &   \num{0.68}  &   --- \\
            KMT2D  &     \SI{17}{\percent}   &   \num{0.74}  &   \textbf{0.76} \\
            BRAF   &     \SI{17}{\percent}   &   \textbf{0.75}  &   \num{0.67} \\
            FBXW7  &     \SI{16}{\percent}   &   \num{0.63}  &   <\num{0.65} \\
            \bottomrule
        \end{tabular}
    \scriptsize
    \caption{Colon (n = 268)}
    \label{tab:coad}
    \end{subtable}
    ~
    \begin{subtable}[t]{0.315\textwidth}
        \scriptsize
        \begin{tabular}{l | r || r | r }
            \toprule
            \multirow{2}{*}{Gene}   & \multicolumn{1}{c||}{\multirow{2}{0.8cm}{\new{Preva-lance}}}    
            &  \multicolumn{1}{c|}{Ours}      &   \multicolumn{1}{c}{Ref.\cite{kather2020pan}}   \\
             & & \multicolumn{1}{c|}{\new{[AUC]}}   & \multicolumn{1}{c}{\new{[AUC]}}   \\
            \midrule
            TP53   &   \SI{50}{\percent}   &   \num{0.71}     &   \textbf{0.72} \\
            KRAS   &   \SI{29}{\percent}   &   \textbf{0.62}  &   <\num{0.6} \\
            FAT4   &   \SI{15}{\percent}   &   \num{0.67}     &   --- \\
            STK11  &   \SI{14}{\percent}   &   \textbf{0.65}  &   \num{0.6} \\
            EGFR   &   \SI{13}{\percent}   &   \textbf{0.70}  &   <\num{0.6} \\
            KMT2C  &   \SI{13}{\percent}   &   \num{0.49}     &   <\num{0.6} \\
            NF1    &   \SI{12}{\percent}   &   \num{0.57}     &   <\num{0.6} \\
            SETBP1 &   \SI{10}{\percent}   &   \num{0.59}     &   <\num{0.6} \\
            \bottomrule
        \end{tabular}
    \scriptsize
    \caption{Lung (n = 461)}
    \label{tab:luad}
    \end{subtable}
    % \scriptsize
    \caption{RoC AUC scores for genetic mutation prediction on TCGA Data.}
    % \vspace{-0.25cm}
    \label{tab:tga_results}
\end{table}

\subsection{Detecting Molecular Alterations}

To gain further insights into the performance of our approach, we address the task of detecting molecular alterations from image features using the datasets discussed in \Cref{sec:TCGA} and compare our results to the study by Kather et al. \cite{kather2020pan}. For our study, we consider the top \num{8} most prevalently mutated genes in each cohort and report the AUC scores in \Cref{tab:tga_results}. Note that this differs from the approach used in \cite{kather2020pan}, who evaluate the prediction performance on a total of \num{95} known cancer driving genes and report the top \num{8} highest scoring results.

\subsubsection{Experimental Setup} 

We employ a patient-level \num{5}-fold cross-validation and use all folds as test-set. No parameter-tuning, thresholds or training decisions are done using those folds. We use the default parameters of our model discussed in \Cref{sec:training} and train the model with these parameters only once on each of the \num{5} folds. In addition, we do not apply any early stopping based on test scores, but train the model for \num{72} epochs and evaluate the scores after the final epoch. We use a multi-label classification approach for this experiment. We train one network per dataset, each with \num{8} binary classification outputs. We apply a softmax-crossentropy loss on each of them and average them (without weights) for training. Note that this approach is different from \cite{kather2020pan} who train a separate network for each gene.

The datasets contain multiple slides for some patients. For training, we choose one slide during each epoch for each patient at random. For inference, we average the confidences over all slides per patient. We perform a patient-level split, i.e. all slides of a patient are part of the same fold.

We compare our results to Kather et al. \cite{kather2020pan}, since the study also performs patient-level cross-validation on their entire cohort.
% We exclude the results from Coudray et al. \cite{coudray2018classification} from our comparison, since they perform a slide-level rather than patient level split. In addition, they evaluate their performance on a custom subset of the data. We believe that this makes it hard to compare the results. 
We note, that our cohort is slightly different from the cohort used in the reference study \cite{kather2020pan} for a number of reasons. Note that Kather et al. manually inspect all slides in the cohort and remove slides of subpar quality. In addition, a number of diagnostic slides have been removed from the TCGA dataset in \num{2021}, due to PII leaking in the images. Lastly, \cite{kather2020pan} uses a custom bioinformatics pipeline to compute the mutation information from the raw NGS data which yields target data for more patients. In summary, the reference study \cite{kather2020pan} uses a larger, higher quality dataset. 

% \section{Conclusion \& Future Work}

\section{Discussion \& Conclusion}
This paper presents a novel $k$-Siamese convolutional neural network architecture for the classification of whole slide images in digital pathology. 
The method is trained end-to-end and does not require auxiliary annotations, which are tedious, time-consuming, and expensive to generate.

In our ablation study, we show that our method is able to clearly outperform commonly used two-stage approaches. We observe that adding a segmentation step to our model only leads to very minor improvement in the AUC score, which proofs \new{that the $k$-Siamese model provides an efficient way of dealing with the label noise issue inherent to tile based processing}.
% This shows that restricting the classifier to the ROI is therefore not required with an efficient architecture as our $k$-Siamese model. 
\new{In addition, our experiments confirm the results shown in \cite{laleh2021benchmarking} that many recently proposed end-to-end methods are unable to outperform the widely used two-stage prediction pipeline. Those methods effectively trade annotation effort for prediction performance. In contrast, our approach is able to deliver state-of-the-art performance without requiring auxiliary annotations.} 

%We believe that the benefit of training on all available clinical data is greater in most applications and avoids extra costs of obtaining auxiliary annotation. We argue that increasing the dataset size is a much more cost effective alternative to improve model performance.

Further experiments on TCGA data reveal that our approach is also highly competitive with the published results by Kather et al. \cite{kather2020pan}: for most genes, our method is able to produce a higher response, painting a clearer picture which mutations have an impact on the morphology of the tumor. In contrast to \cite{kather2020pan}, we are able to produce these results based exclusively on publicly available data, without the need for additional histological annotations. This makes it much easier to reproduce our results, but also allows to explore many more questions and tasks with minimal efforts. 

We hope that the straight-forward implementation of our method, combined with its ability to outperform state-of-the-art approaches, will support further research on the identification of cancer phenotypes by digital pathology and ultimately enable personalized therapies for more patients in future.

\subsubsection{Acknowledgements} 

\new{The research presented in this work was funded by the German Federal Ministry of Education and Research (BMBF) as part of the CancerScout project (13GW0451). We thank all members of the CancerScout Consortium for their contributions, in particular Rico Brendtke and Tessa Rosenthal for organizational and administrative support as well as Sven Winkelmann and Monica Toma for performing various tasks in relation to data privacy, storage and transfer. In addition, we like to thank Christian Marzahl for his support during the installation and adaptation of the EXACT label server. Last but not least, we like to thank Matthias Siebert and Tobias Heckel for insightful discussions about the TCGA Dataset and the associated Omics data.}

\bibliographystyle{splncs04}
\bibliography{miccai}
\end{document}